\documentclass[final, 11pt, 3p, number, sort&compress]{elsarticle}

\usepackage{    amssymb,amsthm,amsmath,tikz,
                        multirow,booktabs,mathrsfs,enumitem,
                        graphicx,dcolumn,bm,algorithm }
\usepackage[hidelinks, colorlinks=true, linkcolor=blue,  anchorcolor=blue, citecolor=blue, filecolor=blue, menucolor=blue, urlcolor=blue]{hyperref}
\usepackage[noend]{algpseudocode} 
\usepackage{subfig}

\usepackage{geometry}
\definecolor{myred}{HTML}{D62728}
\definecolor{myblue}{HTML}{1F77B4}
\newcommand{\redsolidline}{\raisebox{2pt}{\protect\tikz{\protect\draw[-,myred,solid,line width = 1pt](0,0) -- (17pt,0);}}}
\newcommand{\bluesolidline}{\raisebox{2pt}{\protect\tikz{\protect\draw[-,myblue,solid,line width = 1pt](0,0) -- (17pt,0);}}}
\newcommand{\reddashedline}{\raisebox{2pt}{\protect\tikz{\protect\draw[-,myred,dash pattern={on 5pt off 1pt on 5pt off 1pt},line width = 1pt](0,0) -- (17pt,0);}}}
\newcommand{\bluedashedline}{\raisebox{2pt}{\protect\tikz{\protect\draw[-,myblue,dash pattern={on 5pt off 1pt on 5pt off 1pt},line width = 1pt](0,0) -- (17pt,0);}}}

\makeatletter
\def\ps@pprintTitle{%
   \let\@oddhead\@empty
   \let\@evenhead\@empty
   \def\@oddfoot{\footnotesize\itshape
        \hfill \today}
}
\def\els@aparagraph[#1]#2{\elsparagraph[#1]{#2}}
\def\els@bparagraph#1{\elsparagraph*{#1}}
\makeatother

\begin{document}

\begin{frontmatter}

\title{Informative Path Planning for Extreme Anomaly Detection in \\ Environment Exploration and Monitoring} 

\author[]{Antoine Blanchard\corref{cor1}}
\ead{ablancha@mit.edu}
\cortext[cor1]{Corresponding author}
\author[]{Themistoklis Sapsis}

\address{Department of Mechanical Engineering \\ Massachusetts Institute of Technology \\ Cambridge, MA 02139}

\begin{abstract}
An unmanned autonomous vehicle (UAV) is sent on a mission to explore and reconstruct an unknown environment from a series of measurements collected by Bayesian optimization.  The success of the mission is judged by the UAV's ability to faithfully reconstruct any anomalous features present in the environment, with emphasis on the extremes (e.g., extreme topographic depressions or abnormal chemical concentrations).  We show that the criteria commonly used for determining which locations the UAV should visit are ill-suited for this task.  We introduce a number of novel criteria that guide the UAV towards regions of strong anomalies by leveraging previously collected information in a mathematically elegant and computationally tractable manner.  We demonstrate superiority of the proposed approach in several applications, including reconstruction of seafloor topography from real-world bathymetry data, as well as tracking of dynamic anomalies. A particularly attractive property of our approach is its ability to overcome adversarial conditions, that is, situations in which prior beliefs about the locations of the extremes are imprecise or erroneous.
\end{abstract}

\end{frontmatter}

\section{Introduction}
\label{sec:1}

With the rise of automation and artificial intelligence, a growing number of tasks deemed too tedious or too perilous for humans have been delegated to unmanned autonomous vehicles (UAV).  This includes missions related to environment exploration and monitoring in which an UAV is tasked with producing a map for a quantity of interest (e.g., pollutant concentration, terrain elevation, or vegetation growth) by collecting measurements at various locations across a region of interest (e.g., a reservoir, a city, or a crop) \cite{singh2010modeling,dunbabin2012robots,marchant2012bayesian,hitz2017adaptive,flaspohler2018near}.  The data collected by the UAV can be used to construct a statistical model for the quantity of interest, which in turn can be used for analysis and policy making.  Of course, the statistical model is only as good as the measurements made by the UAV.  Therefore, the question of data collection (i.e., how, when, and where to make measurements) is of paramount importance, especially from the standpoint of detecting \emph{anomalies} in the environment.

Path-planning algorithms for environment exploration come in two flavors.  Approaches in which the UAV decides on its next move one step at a time are referred to as \emph{myopic} \cite{stachniss2005information,marchant2014bayesian}.  Myopic algorithms are suitable for most situations but lack a mechanism for anticipation, which may be problematic in cases where path-planning decisions may have negative long-term consequences (e.g., the UAV gets stuck because of maneuverability constraints).  On the other hand, \emph{non-myopic} algorithms operate on sequences of destinations, which allows them to look further into the future \cite{meliou2007nonmyopic,singh2009nonmyopic,morere2016bayesian,morere2017sequential}.  The main tool for this is the \textit{partially observable Markov decision process}, which assigns a reward to each admissible sequence of actions.  Non-myopic approaches are computationally complex and incredibly expensive, which is why myopic approaches are often preferred---including in the present work.  We note, however, that the ideas presented here can be extended to the non-myopic setting notwithstanding the higher computational complexity.

The flagship feature of myopic algorithms is that they naturally lend themselves to Bayesian optimization \cite{brochu2010tutorial,shahriari2015taking}, allowing the UAV to a) incorporate prior belief about the environment, and b) decide on its next move by compromising between exploration of the space and exploitation of the available information.  At the heart of Bayesian optimization is the acquisition function, which guides the UAV throughout the mission.  Many acquisition functions have been proposed for environment exploration \cite{martinez2009bayesian,marchant2012bayesian,marchant2014bayesian,bai2016information,morere2017sequential} but they all have one major flaw, namely, they have no robust mechanism to identify anomalous environmental features.  For example, the approach of \cite{marchant2012bayesian} requires the user to specify the values of two ad hoc parameters which must be tuned on a case-by-case basis, with no foolproof guidelines on how to do so.  

The main contribution of this work is the introduction of two novel acquisition functions that are specifically designed for anomaly detection in environment exploration.  The proposed acquisition functions have the following advantages:
\begin{enumerate}
\item They are based on a probabilistic treatment of what constitutes an anomaly, thereby eliminating the need for ad-hoc parameters;
\item Their computational complexity is comparable with that of traditional acquisition functions, making them suitable for online path planning and monitoring;
\item They provide the UAV operator with a mechanism to instill any prior beliefs they may have about the locations of anomalies while allowing the UAV to \emph{correct and refine} the operator beliefs ``on the fly'' as more information is collected. 
\end{enumerate}

The remainder of the paper is structured as follows.  We present the problem and approach in Section \ref{sec:2}, introduce the acquisition functions in Section \ref{sec:3}, evaluate their performance in Section \ref{sec:4}, and offer some conclusions in Section \ref{sec:5}.

\section{Problem Formulation and Approach}
\label{sec:2}

\subsection{Formulation of the Problem}
\label{sec:21}

We consider the problem of environment exploration in which an UAV is tasked with reconstructing the spatiotemporal distribution of a quantity of interest  $f : \mathbb{R}^2 \times \mathbb{R}^+ \longrightarrow \mathbb{R}$ over a region of interest $\mathcal{Z} \subset \mathbb{R}^2$ and a time interval of interest $[0, T] \subset \mathbb{R}^+$.  (We assume that $f$ is Lipschitz continuous and $\mathcal{Z}$ is compact.)  To reconstruct the map $f$, the UAV is allowed to explore the space $\mathcal{Z}$ and collect measurements at locations it deems informative.  Uncertainty in observations is modeled with additive Gaussian noise, so that each measurement made by the UAV can be written as 
\begin{equation}
y = f(\mathbf{z}, t) + \varepsilon, \quad \varepsilon \sim \mathcal{N}(0, \sigma_n^2),
\label{eq:1}
\end{equation}
where $\mathbf{z} = [z_1,z_2]^\mathsf{T}$ is a vector of coordinates identifying the UAV's position in the Euclidean plane, and $t$ is the time variable.  The combination of position $\mathbf{z}$ and orientation $\theta$ (with $\theta=0$ taken to coincide with the $z_1$-axis) completely determines the \emph{pose} of the UAV.  In what follows we will find it useful sometimes to view $f$ as a function of a single input vector $\mathbf{x} = \{ \mathbf{z}, t\}$.

The two key issues in environment reconstruction are data acquisition (i.e., when and where to collect measurements) and environment modeling (i.e., how to leverage measurements to construct an accurate model of the environment).  From a modeling perspective, the challenge is to predict the value of a process that depends on space \textit{and} time given a limited number of measurement points.  Notable approaches for spatiotemporal modeling include space--time process convolutions \cite{higdon2002space} and kernel extrapolation methods for distribution mapping algorithms \cite{reggente2009using}, but the most popular of all is arguably Gaussian process (GP) regression, which we briefly review in Section \ref{sec:22}.  From a data-acquisition perspective, the challenge is to identify the locations that provide the most information about the underlying environmental process.  This question, central to the present work, is discussed in Section \ref{sec:23} and the subsequent sections.

\subsection{Gaussian Process Regression for Environment Modeling}
\label{sec:22}

To reconstruct the latent function $f$ from a limited number of noisy measurements, a natural solution is to use a surrogate model, which we denote by $\bar{f}$.  In this work we use a Bayesian approach based on Gaussian process (GP) regression \cite{rasmussen2006gaussian}.  In the context of environment reconstruction, GP regression has several advantages.  First, Gaussian processes are agnostic to the internal intricacies of the unknown map $f$.  Second, they provide a way to quantify uncertainty associated with noisy observations.  Third, they are equipped with a mechanism that takes into account possible correlations across space and time, allowing them to deliver good performance on spatially-correlated data \cite{singh2010modeling,marchant2014bayesian}.  Fourth, they are robust, versatile, easy to implement, and inexpensive to train when the input dimension and number of measurement points are not too large.  For spatiotemporal processes the input dimension never exceeds four, i.e., at most three spatial dimensions and one temporal dimension.

For a dataset $\mathcal{D}=\{\mathbf{X}, \mathbf{y}\}$ of input--output pairs and a zero-mean GP with covariance function $k(\mathbf{x},\mathbf{x}')$, the random process $\bar{f}(\mathbf{x})$ conditioned on $\mathcal{D}$ follows a normal distribution with posterior mean and variance 
\begin{subequations}
\begin{gather}
\mu(\mathbf{x}) = k(\mathbf{x}, \mathbf{X}) \mathbf{K}^{-1} \mathbf{y}, \label{eq:2a}\\
\sigma^2(\mathbf{x}) = k(\mathbf{x},\mathbf{x}) - k(\mathbf{x}, \mathbf{X}) \mathbf{K}^{-1} k(\mathbf{X},\mathbf{x}),\label{eq:2b}
\end{gather}
\end{subequations}
respectively, where $\mathbf{K} = k(\mathbf{X},\mathbf{X})+ \sigma_n^2 \mathbf{I}$.  The posterior mean can be used to predict the value of the quantity of interest at any point $\mathbf{x}$, and the posterior variance to quantify uncertainty in prediction at that point.  As discussed in Section \ref{sec:21}, here the input variable $\mathbf{x}$ belongs to an augmented space which is constructed by appending the time variable $t$ to the physical variables $\mathbf{z}$.  Consequently, GP regression allows to infer the value of the latent function $f$ at any spatial location, but also at any point in time, past or future.

In GP regression, the covariance function is the main building block for practitioners to encode structure (e.g., symmetry or invariance) in the model.  For time-dependent environments, it is common to distinguish between separable and non-separable covariance functions \cite{marchant2014bayesian}.  The former take the general form $k(\mathbf{x},\mathbf{x}') = k(\mathbf{z},\mathbf{z}') k(t,t')$ and are useful when the spatial variables are decoupled from the temporal variable in the latent function \cite{marchant2014bayesian}.  In general, however, the coupling between space and time in the environment is complex or unknown, and as a result it is often preferable to use a non-separable covariance function.  In this work we use a radial-basis-function (RBF) kernel with automatic relevance determination,
\begin{equation}
k(\mathbf{x},\mathbf{x}') = \sigma_f^2 \exp [ -(\mathbf{x} - \mathbf{x}')^\mathsf{T} \mathbf{\Theta}^{-1}(\mathbf{x} - \mathbf{x}') /2],
\label{eq:3}
\end{equation}
where $\mathbf{\Theta}$ is a diagonal matrix containing the lengthscales for each dimension and $\sigma_f^2$ is a scaling parameter.  For a given dataset, the hyper-parameters $\{ \sigma_f^2, \mathbf{\Theta}\}$ are trained by maximum likelihood estimation \cite{rasmussen2006gaussian}.  


\subsection{Bayesian Optimization for Informative Path Planning}
\label{sec:23}

The next question is to design an optimal strategy for data acquisition.  As discussed in Section \ref{sec:21}, the challenge is to select sensing locations in such a way that the resulting surrogate model is of sufficiently high fidelity across the region and time interval of interest.  One simple strategy is for the UAV to visit a large number of precomputed locations that densely cover the search space, which is essentially equivalent to having a fixed network of sensors \cite{guestrin2005near}.  In practice, however, this approach is intractable because exploration missions are always done on a budget with limited time and resources.

A better strategy is for the UAV to proceed sequentially and decide on its next destination based on the information it has collected so far.  This information can be leveraged to improve the surrogate model $\bar{f}$ in real time.  At each iteration the next destination is selected by minimizing an acquisition function $a : \mathbb{R}^2 \times \mathbb{R}^+ \longrightarrow \mathbb{R}$ which guides the UAV in its exploration of the space.  Once the allocated budget is exhausted, the mission is terminated and the surrogate model constructed by the UAV can be used in analyses as a substitute for the unknown map $f$ (Algorithm \ref{alg:1}).  This algorithm is at the foundation of Bayesian optimization \cite{krause2011contextual,shahriari2015taking}.  Its success is conditioned on two key components: a) the surrogate model $\bar{f}$, which encapsulates the UAV's belief about what the environment looks like given the data it has collected; and b) the acquisition function $a$, upon which the UAV relies to plan its next move.

\begin{algorithm}
   \caption{Bayesian optimization for the ``next-best-view'' problem.}
   \label{alg:1}
\begin{algorithmic}[1]
   \State {\bfseries Input:} Mission duration $T$, initial UAV position $\mathbf{z}_0$ and orientation $\theta_0$, initial dataset $\mathcal{D} = \{\mathbf{x}_i, y_i\}_{i=1}^{n_\mathit{init}}$
   \State {\bfseries Initialize:} Surrogate model $\bar{f}$ trained on $\mathcal{D}$
   \While{$t \leq T$}
      \State Select next destination as $\mathbf{z}_{n+1} = \operatorname*{arg\,min}_{\mathbf{z} \in \mathcal{Z}} a(\mathbf{z}, t; \bar{f}, \mathcal{D})$ \label{lst:line4}
        \State Record measurement $y_{n+1}$ at $\mathbf{x}_{n+1} = \{\mathbf{z}_{n+1}, t_{n+1}\}$              \State Augment dataset: $\mathcal{D} \leftarrow \mathcal{D} \cup \{\mathbf{x}_{n+1} , y_{n+1} \}$
   \State Update surrogate model
   \EndWhile
   \Return Final surrogate model $\bar{f}$
\end{algorithmic}
\end{algorithm}

In practice the sampling frequency of the UAV sensors is much higher than the frequency of the decision making, and as a result the approach in Algorithm \ref{alg:1} is suboptimal because measurements are only collected at destination \cite{morere2017sequential}.  Therefore, to replicate the conditions of an actual field experiment, we modify Algorithm \ref{alg:1} in two ways.  First, we allow the UAV to collect measurements periodically (with sampling period $t_s$) as it travels from $\mathbf{z}_{n}$ to $\mathbf{z}_{n+1}$.  Second, to make the most of this additional data, we change the path-planning policy in Line \ref{lst:line4} of Algorithm \ref{alg:1} to
\begin{equation}
\mathbf{z}_{n+1} = \operatorname*{arg\,min}_{\mathbf{z} \in \mathcal{Z}} \int_{\mathcal{S}(\mathbf{z}_n, \mathbf{z})} a(\mathbf{z}, t; \bar{f}, \mathcal{D}_n) \, \mathrm{d}s,
\label{eq:14}
\end{equation}
where $\mathcal{S}(\mathbf{z}_n, \mathbf{z})$ denotes the path taken by the UAV to reach candidate destination $\mathbf{z}$ from its current position $\mathbf{z}_n$.  This approach, known as \emph{informative path planning} \cite{morere2017sequential}, is summarized in Algorithm \ref{alg:2}.

\begin{algorithm}
   \caption{Bayesian optimization for informative path planning.}
   \label{alg:2}
\begin{algorithmic}[1]
   \State {\bfseries Input:} Mission duration $T$, initial UAV position $\mathbf{z}_0$ and orientation $\theta_0$, initial dataset $\mathcal{D} = \{\mathbf{x}_i, y_i\}_{i=1}^{n_\mathit{init}}$
   \State {\bfseries Initialize:} Surrogate model $\bar{f}$ trained on $\mathcal{D}$
   \While{$t \leq T$}
      \State Select next destination as in \eqref{eq:14}
    \While{traveling from $\mathbf{z}_n$ to $\mathbf{z}_{n+1}$}
        \If{$\mod(t, t_s) = 0$}
        \State Record measurement $y_t$ at $\mathbf{x}_t = \{\mathbf{z}_t, t\}$
        \State Augment dataset: $\mathcal{D} \leftarrow \mathcal{D} \cup \{\mathbf{x}_t , y_t \}$
        \EndIf
   \EndWhile
   \State Update surrogate model
   \EndWhile
   \Return Final surrogate model $\bar{f}$
\end{algorithmic}
\end{algorithm}

The primary difference between the traditional ``next-best-view'' problem (Algorithm \ref{alg:1}) and  the informative path-planning approach (Algorithm \ref{alg:2}) is that the latter takes into account the UAV path in the calculation of the next destination.  This allows the UAV to optimize its motion not only based on the expected value of the latent function at destination, but also based on the quality of the information the UAV is expecting to collect during its journey from $\mathbf{z}_n$ to $\mathbf{z}_{n+1}$.  

Typically the path $\mathcal{S}(\mathbf{z}_n, \mathbf{z})$ is a parametric curve satisfying certain constraints related to continuity, smoothness, and curvature.  For a given candidate destination $\mathbf{z}$, there is generally an infinite number of paths connecting $\mathbf{z}_n$ and $\mathbf{z}$, so in practice the path parametrization is either specified in advance in a way that ensures uniqueness or optimized on the fly using an additional layer of optimization as in Reference \cite{marchant2014bayesian}.  In this work consecutive destinations are connected by a Dubins path with fixed turning radius along which the UAV travels at constant speed.   Dubins paths are popular in robotics and path planning owing to their simple geometric construction and computational tractability \cite{dubins1957curves}.   This parametrization ensures that the UAV path is $C^1$-continuous.  

As it stands, the path-planning policy \eqref{eq:14} gives an unfair advantage to shorter routes over longer ones.  A possible remedy is to normalize the integral in \eqref{eq:14} by the path length.  A more practice-driven approach is to restrict the set of admissible destinations to a subset $\mathcal{A}_n \subset \mathcal{Z}$, mimicking the fact that in practice the UAV has limited field of view and limited sensor range.  This is the approach we will use in this work.  We also note that for most acquisition functions, the integral in \eqref{eq:14} is not analytic; in this work, we evaluate it using the trapezoidal rule.

\subsection{Acquisition Functions for Environment Exploration}
\label{eq:23}

We now address the question of data acquisition for environment reconstruction.  The acquisition functions introduced below can be used ``as is'' in Algorithm \ref{alg:1}, or as the integrand in \eqref{eq:14} for use in Algorithm \ref{alg:2}. 

In environment exploration, the role of the acquisition function is to favor exploration of regions where uncertainty is high.  We briefly review two common approaches to achieving this.

\paragraph{Uncertainty sampling (US):}  The most intuitive approach is for the UAV to go where the predictive variance of the GP model is the largest \cite{mackay1992information}, that is,
\begin{equation}
a_\textit{US}(\mathbf{x}) = \sigma^2(\mathbf{x}).
\label{eq:4}
\end{equation}
This approach is tantamount to minimizing the mean squared error between $\bar{f}$ and $f$ \cite{kleijnen2004application,beck2016sequential}, and therefore ensures that model uncertainty is distributed somewhat evenly over the search space. 

\paragraph{Integrated variance reduction (IVR):}  Another approach is to consider the effect of observing a hypothetical ``ghost'' point $\mathbf{x}$ on the overall model variance \cite{cohn1994neural}.  This effect is measured by 
\begin{equation}
a_\textit{IVR}(\mathbf{x}) = \int \left[ \sigma^2(\mathbf{x}') - \sigma^2(\mathbf{x}'; \mathbf{x}) \right] \mathrm{d}\mathbf{x}' = \frac{1}{\sigma^2(\mathbf{x})} \int \mathrm{cov}^2(\mathbf{x}, \mathbf{x}') \, \mathrm{d}\mathbf{x}', 
\label{eq:5}
\end{equation}
where $\sigma^2(\mathbf{x}'; \mathbf{x})$ is the predictive variance at $\mathbf{x}'$ had $\mathbf{x}$ been observed, and $\mathrm{cov}^2(\mathbf{x}, \mathbf{x}')$ is the posterior covariance between $\mathbf{x}$ and $\mathbf{x}'$ \cite{gramacy2009adaptive,blanchard2020output}. Therefore, maximizing IVR has the effect of maximally reducing the overall model variance.  \\

If the UAV operator wants to focus exploration on certain regions of the search space, or if they have prior beliefs about where relevant environmental features might be located, then it is possible to bias exploration by incorporating a prior $p_\mathbf{x}(\mathbf{x})$ over the input space in US and IVR.  The prior acts as a sampling weight, leading to
\begin{gather}
a_\textit{US-IW}(\mathbf{x}) = \sigma^2(\mathbf{x}) p_\mathbf{x}(\mathbf{x}), \label{eq:6} \\
a_\textit{IVR-IW}(\mathbf{x}) = \frac{1}{\sigma^2(\mathbf{x})} \int \mathrm{cov}^2(\mathbf{x}, \mathbf{x}') p_\mathbf{x}(\mathbf{x}') \, \mathrm{d}\mathbf{x}',\label{eq:7}%
\end{gather}
where the suffix ``IW'' stands for ``input-weighted''.  If $\mathbf{x}$ carries contextual information in addition to spatial information, then the input prior can be decomposed as $p_\mathbf{x}(\mathbf{x}) = p_\mathbf{z}(\mathbf{z}) p_t(t)$, with the temporal prior $p_t$ typically being uniform.  If no prior knowledge is available, then $p_\mathbf{z}$ too can be chosen uniform, in which case US-IW and IVR-IW reduce to US and IVR, respectively.

The idea of using a prior as a sampling weight was first suggested by Sacks et al. \cite{sacks1989design} in the context of sequential design of computer experiments. To the best of our knowledge, it has never been applied to problems related to environment exploration as a way of emphasizing certain regions of the input space.  The closest instance of which we are aware is the work of Oliveira et al. \cite{oliveira2019bayesian,oliveira2020bayesian} in which a prior is placed on the input space in order to account for \emph{localization noise}, i.e., the error in estimating the UAV position resulting from imperfections in the UAV sensors, actuators, and motion controllers.  This is quite different from the proposed approach in which the UAV position is assumed to be known with exactitude and the input prior is used as a mechanism to highlight certain regions of the search space before the mission starts. \\

While other acquisition functions exist, either they are computationally more complex or they have narrower applicability than those previously discussed.  For example, the mutual information, in general, cannot be written in closed form and thus loses out to US, the latter being a good approximation for the former \cite{mackay1992information,chaloner1995bayesian}.  Likewise, the upper confidence bound (UCB), the expected improvement (EI), and the probability of improvement (PI) are not appropriate for environment reconstruction because they are designed for optimization, not exploration \cite{jones1998efficient,srinivas2009gaussian}.  And while UCB, EI and PI each contain an ad-hoc parameter that controls the trade-off between exploration and exploitation, setting the value of that parameter to a large number in order to favor exploration is moot because in that limit UCB, EI and PI are equivalent to US (see Section S1 in the Supplementary Material).  Variants of these criteria \cite{verdinelli1992bayesian,lam2008sequential,marchant2012bayesian,morere2017sequential} suffer from the same shortcomings.

\section{Methods}
\label{sec:3}

In this section, we introduce two novel acquisition functions for reconstruction of anomalous environment.  Both share three critical features: a) they leverage information collected previously by the UAV and assign more weight to regions of the search space where the map $f$ is thought to exhibit strong anomalies;  b) they allow incorporation of a prior $p_\mathbf{x}(\mathbf{x})$ over the search space, with expectations about potential benefits being similar to those discussed earlier for US-IW and IVR-IW; and c) their computational complexity is comparable to that of traditional acquisition functions.  

\subsection{Output-Informed Acquisition Functions for Anomalous Environment}

We begin with a definition of what constitutes an anomalous environment.  A map $f : \mathcal{X} \longrightarrow \mathbb{R}$ is \emph{anomalous} if the conditional probability density function (pdf) of the output $p_{f|\mathbf{x}}$ is heavy-tailed. A pdf is heavy-tailed when at least one of its tails is not exponentially bounded.  Heavy tails are the manifestation of high-impact events occurring with low probability, and are therefore appropriate to characterize anomalies in output values.  Heavy-tailed distributions commonly arise in the study of risk \cite{embrechts2013modelling} and extreme events \cite{albeverio2006extreme} but as far as we know they have not been considered in the context of environment exploration.  In what follows, we drop the conditional notation for clarity.  

The proposed definition suggests a strategy for the UAV to decide on its next destination.  At each iteration, the UAV can use the pdf of the GP mean $p_{\mu}$ as a proxy for $p_{f}$ and select the next destination so that uncertainty in $p_{\mu}$ is most reduced.  The latter can be quantified by
\begin{equation}
a_L(\mathbf{x}) = \int \left| \log p_{\mu_+}(y) - \log p_{\mu_-}(y)  \right| \mathrm{d}y,
\label{eq:8}
\end{equation}
where $\mu_{\pm}(\mathbf{x}'; \mathbf{x})$ denotes the upper and lower confidence bounds at $\mathbf{x}'$ had the data point $\{\mathbf{x}, \mu(\mathbf{x})\}$ been collected, that is, $\mu_{\pm}(\mathbf{x}'; \mathbf{x}) =  \mu(\mathbf{x}') \pm \sigma^2(\mathbf{x}'; \mathbf{x})$.  The use of logarithms in \eqref{eq:8} places extra emphasis on the pdf tails in which critical information about abnormal features is encapsulated.  

The above metric enjoys attractive convergence properties \cite{mohamad2018sequential} but is cumbersome to compute (let alone minimize) and therefore unsuitable for online path planning.  To combat this, we rely on the property that $a_L(\mathbf{x})$ is bounded above (up to a multiplicative constant \cite{sapsis2020output}) by 
\begin{equation}
a_B(\mathbf{x}) = \int \sigma^2(\mathbf{x}'; \mathbf{x})  \frac{p_\mathbf{x}(\mathbf{x}')}{p_\mu(\mu(\mathbf{x}'))} \, \mathrm{d}\mathbf{x}'.
\label{eq:9}
\end{equation}
Equation \eqref{eq:9} is a massive improvement over \eqref{eq:8} from the standpoint of reducing complexity.  More importantly, it reveals an unexpected connection between the metric $a_L$ (whose primary focus is the reduction of uncertainty in pdf tails) and the IVR-type acquisition functions of Section \ref{eq:23}.  Indeed, it only takes a few lines to show that $a_B(\mathbf{x})$ is strictly equivalent to 
\begin{equation}
a_\textit{IVR-LW}(\mathbf{x}) = \frac{1}{\sigma^2(\mathbf{x})} \int \mathrm{cov}^2(\mathbf{x}, \mathbf{x}') \frac{p_\mathbf{x}(\mathbf{x}')}{p_\mu(\mu(\mathbf{x}'))} \, \mathrm{d}\mathbf{x}',
\label{eq:10}
\end{equation}
which is clearly related to \eqref{eq:5} and \eqref{eq:7}, with the ratio $p_\mathbf{x}(\mathbf{x})/p_\mu(\mu(\mathbf{x}))$ playing the role of a sampling weight.  By the same logic, we introduce
\begin{equation}
a_\textit{US-LW}(\mathbf{x}) = \sigma^2(\mathbf{x}) \frac{p_\mathbf{x}(\mathbf{x})}{p_\mu(\mu(\mathbf{x}))}
\label{eq:11}
\end{equation}
as the ``likelihood-weighted'' (LW) counterpart of US and US-IW.  For details about the derivation, we refer the reader to Blanchard and Sapsis \cite{blanchard2020output,blanchard2021bayesian}.

\subsection{The Likelihood Ratio and its Benefits}
In the importance-sampling literature, the ratio
\begin{equation}
w(\mathbf{x}) = \frac{p_\mathbf{x}(\mathbf{x})}{p_\mu(\mu(\mathbf{x}))}
\label{eq:12}
\end{equation}
is referred to as the \emph{likelihood ratio} \cite{owen2013monte}.  The likelihood ratio is important in cases where some points are more important than others in determining the value of the output.  When used in an acquisition function, it acts as a sampling weight, assigning to each point $\mathbf{x} \in \mathcal{X}$ a measure of \emph{relevance} defined in probabilistic terms.  For points with equal probability of being observed ``in the wild'' (i.e., same $p_\mathbf{x}$), the likelihood ratio assigns more weight to those that have a large impact on the magnitude of output (i.e., small $p_\mu$).  For points with equal impact on the output (i.e., same $p_\mu$), it promotes those with higher probability of occurrence (i.e., large $p_\mathbf{x}$).  In other words, the likelihood ratio favors points associated with \textit{abnormal} output values over points associated with frequent, average output values.  

In environment exploration, the likelihood ratio can be beneficial in two ways.  First, it incorporates field information through the GP mean $\mu(\mathbf{x})$ in such a way that a) no additional ad-hoc parameter is introduced; b) anomalous regions are naturally accentuated due to the probabilistic dependence on the density $p_\mu$; and c) it does not discriminate between abnormally small and abnormally small output values, which is a tremendous advantage over approaches based on PI, EI or UCB.  Second, the likelihood ratio preserves the possibility for the operator to instill prior belief through the density $p_\mathbf{x}$.  But the fact that the input prior is weighted by the output density $p_\mu$ enables the UAV to \emph{correct and refine} the operator beliefs on the fly as more information is collected.  Thus, the likelihood ratio provides a mechanism for the UAV to strike an informed balance between its own representation of the environment and the operator guidelines.  We will see that this feature significantly improves performance in \textit{adversarial} situations, that is, situations where the operator beliefs are imprecise or erroneous.

We should not lose sight of the fact that these benefits may be neutralized if the likelihood ratio is not tractable computationally.  Fortunately, that is not the case.  To see this, we note that to evaluate $w(\mathbf{x})$, we must estimate the pdf of the posterior mean $p_\mu$, typically at each iteration.  This can be done by computing $\mu(\mathbf{x})$ for a large number of input points and applying kernel density estimation (KDE) to the resulting samples.   Fortunately, KDE is to be performed in the (one-dimensional) output space, allowing use of fast FFT-based algorithms which scale linearly with the number of samples \cite{fan1994fast}.  

To evaluate \eqref{eq:10} without resorting to Monte Carlo integration, we follow Blanchard and Sapsis \cite{blanchard2020output,blanchard2021bayesian} and approximate the likelihood ratio with a Gaussian mixture model (GMM):
\begin{equation}
w(\mathbf{x})  \approx \sum_{i=1}^{n_\textit{GMM}} \alpha_i \, \mathcal{N}(\mathbf{x}; \boldsymbol{\omega}_i, \mathbf{\Sigma}_i).
\label{eq:13}
\end{equation}
When combined with the RBF kernel, the GMM approximation renders the integral in \eqref{eq:10} analytic \cite{blanchard2020output,blanchard2021bayesian}.  The number of Gaussian mixtures in \eqref{eq:13} can be kept constant throughout the mission or modified in real time, either according to a predefined schedule or by minimizing the Akaike information criterion (AIC) or the Bayesian information criterion (BIC) at each iteration.

\begin{figure}[!ht]
\centering 
\includegraphics[width=5.5in]{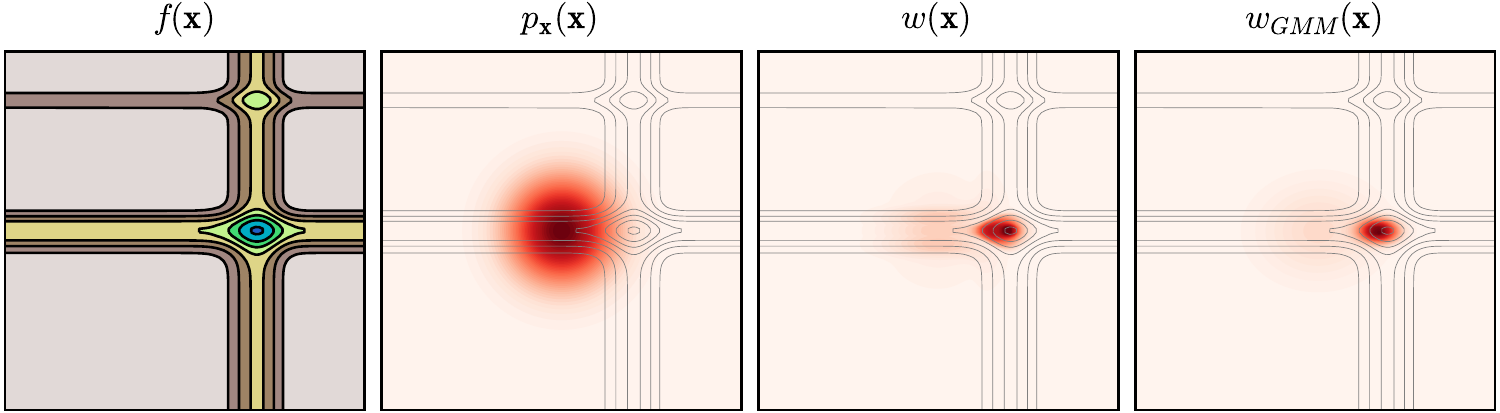} 
\caption{From left to right: Contour plots of the map $f(\mathbf{x})$, the input prior $p_\mathbf{x}(\mathbf{x}) = p_\mathbf{z}(\mathbf{z})$, the likelihood ratio $w(\mathbf{x})$, and the GMM approximation of the likelihood ratio $w_\textit{GMM}(\mathbf{x})$ with two Gaussian mixtures.}
\label{fig:1}
\end{figure}

For an illustration of the benefits provided by the likelihood ratio, we consider the (static) Michalewicz function (S12d) with the input space $[0,\pi]^2$ rescaled to the unit square and a Gaussian prior $p_\mathbf{z}(\mathbf{z}) = \mathcal{N}(\mathbf{0}+1/2, 0.01\mathbf{I})$ in space (see Figure \ref{fig:1}).  The Michalewicz function is characterized by large regions of ``flatland'' interrupted by steep valleys and ridges, with its deepest portion accounting for a tiny fraction of the search space.  For this function, Figure \ref{fig:1} makes it visually clear that the likelihood ratio gives more emphasis to the area where the quantity of interest $f(\mathbf{x})$ assumes abnormally small values, taking precedence over the operator belief that the UAV should focus solely on the center region.  Figure \ref{fig:1} also shows that $w(\mathbf{x})$ can be approximated satisfactorily with a small number of Gaussian mixtures, a key prerequisite for algorithm efficiency.

\section{Results}
\label{sec:4}

\subsection{Experimental Protocol}
\label{sec:41}

\begin{figure}[!ht]
\centering
\includegraphics[width=2.5in]{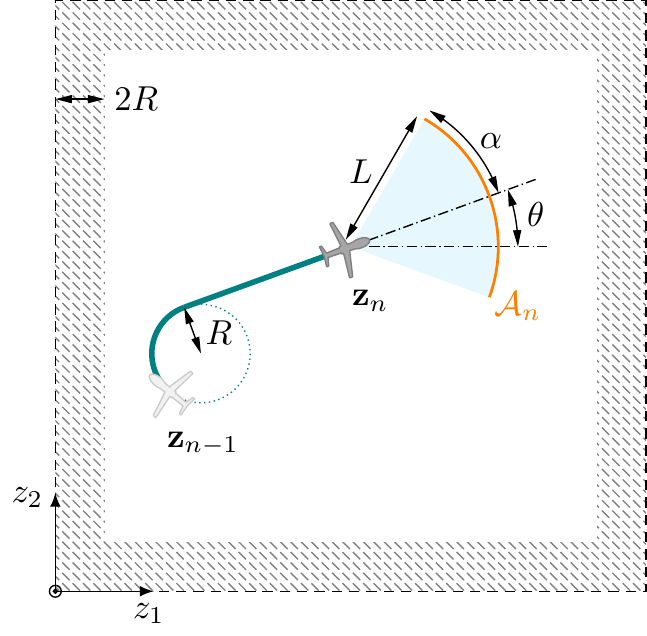} 
\caption{Schematic of the experimental set-up for environment exploration (not to scale).}
\label{fig:fig2}
\end{figure}

To solidify the utility of the likelihood ratio in environment exploration, we perform a series of numerical experiments as per the following protocol.  For each example considered, the search space is rescaled to the unit square $[0,1]^2$ and the UAV initial pose is specified as $\mathbf{z}_0=\mathbf{0}$ and $\theta_0=\pi/4$.  At each iteration, the UAV admissible destinations are constrained to lie on a circular arc with radius $L$ centered at $\mathbf{z}_n$ and subtending an angle $2 \alpha$ which bisects $\theta$.  The parameters $L$ and $\alpha$ characterize the \emph{lookahead distance} and \emph{field of view} of the UAV, respectively.  To avoid situations in which the UAV might venture outside of the search space, we discard from the set of admissible destinations those lying a distance $2 R$ or less from the boundaries of the unit square.  Each mission takes place over the interval $t \in [0,15]$, with the UAV traveling at unit speed and collecting measurements every $t_s=1/15$ time units.  A schematic of the experimental set-up is shown in Figure \ref{fig:fig2}.

To assess performance of the algorithm, we compute the following metrics.
\begin{subequations}
\paragraph{Root mean square error:}
\begin{equation}
\text{rmse}(t) = \sqrt{\frac{1}{N} \sum_{i=1}^N [ f(\mathbf{z}_i, t) - \mu_t(\mathbf{z}_i, t) ]^2},
\end{equation}
where $\{\mathbf{z}_i\}_{i=1}^N$ is a set of $10^5$ samples uniformly distributed over the physical domain, and $\mu_t$ denotes the posterior mean of the GP model in use by the UAV at time $t$. 

\paragraph{Log-pdf error:}
\begin{equation}
\text{pdfe}(t) =  \int  \left| \log p_{f(\mathbf{z}, t)}(y) - \log p_{\mu_t(\mathbf{z}, t)}(y) \right| \, \mathrm{d} y,
\end{equation}
where the densities are estimated using the same $10^{5}$ samples as for the root mean square error.

\paragraph{Distance to minimizer:}
\begin{equation}
\ell(t) =  \Vert \mathbf{z}_t^* - \mathbf{z}^+_t \Vert^2,
\end{equation}
where $\mathbf{z}_t^*$ and $\mathbf{z}^+_t$ are the minimizers for the true map $f(\mathbf{z},t)$ and the GP mean $\mu_t(\mathbf{z},t)$, respectively.

\paragraph{Simple regret:}
\begin{equation}
r(t) =  f(\mathbf{z}^*_t, t) - f(\mathbf{z}_t^+, t).
\end{equation}
\end{subequations}

The root mean square error measures the overall goodness of the GP model with no consideration for anomalies of any kind.  In contrast, the log-pdf error judges the model by its ability to reconstruct the tails of the output pdf, which is where abnormal features ``live''.  The metrics $\ell$ and $r$ quantify the model ability to predict the locations and output value, respectively, of the map global minimizers.  These two metrics are appropriate because in the examples considered the global minimizers are associated with abnormally small output values.

On the first iteration, any admissible destination is as good as any other from the UAV's standpoint because there is only one measurement available (collected at $\mathbf{z}_0$).  To disambiguate the situation, the first destination $\mathbf{z}_1$ is drawn uniformly from the set of admissible destinations $\mathcal{A}_0$.  This introduces an element of randomness in the problem, which is averaged out by repeating the mission many times, each time with a different choice of $\mathbf{z}_1$.  For each example considered, we repeat the mission 50 times, and report the median of the cumulative minimum for the four metrics introduced above.  The error bands indicate a quarter of the median absolute deviation. Our code is available on GitHub\footnote{\url{https://github.com/ablancha/gppath}}.

\subsection{Benchmark Results for Static Test Functions}
\label{sec:42}

\begin{figure}[!ht]
\centering 
\subfloat[Ackley]{\includegraphics[width=1.2in]{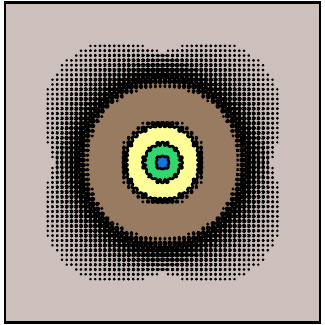}}  ~
\subfloat[Bird]{\includegraphics[width=1.2in]{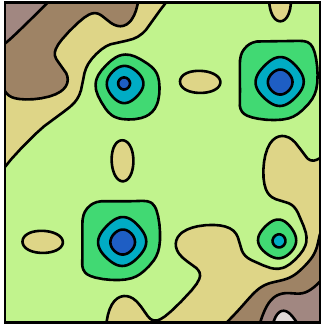}} ~
\subfloat[Bukin06]{\includegraphics[width=1.2in]{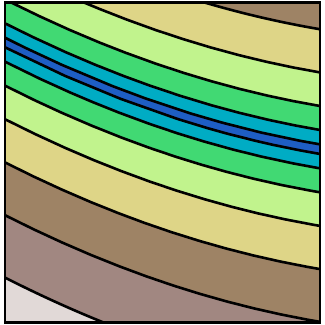}}  ~
\subfloat[Michalewicz]{\includegraphics[width=1.2in]{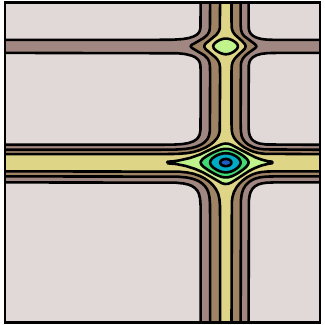}}  ~
\subfloat[Mod. Rosenbrock]{\includegraphics[width=1.2in]{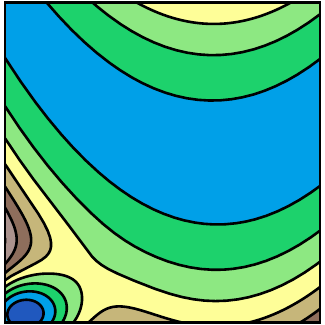}}

\caption{Contour plots of the test functions considered in Section \ref{sec:42}.}
\label{fig:3}
\end{figure}

We evaluate the performance of the proposed criteria on five static test functions commonly used in optimization and uncertainty quantification: Ackley, Bird, Bukin06, Michalewicz, and Modified Rosenbrock.  Analytical expressions are given in Section S2 of the Supplementary Material. These functions were chosen because they are representative of what an anomalous environment may look like in real life.  For example, the steep ridge of the Bukin06 function is reminiscent of an unusually deep oceanic trench; and the Ackley function is evocative of a substance diffusing away from an abnormally potent source (Figure \ref{fig:3}).

In each case, the noise variance is specified as $\sigma_n^2=10^{-3}$ and appropriately rescaled to account for the variance of the map $f$.   We do not set the parameter $\sigma_n^2$ beforehand in the GP model; instead, we let the UAV learn it from data.  We use $L=0.2$, $\alpha=3\pi/4$, and $R=0.02$ for the path-planning algorithm, and $n_\textit{GMM}=2$ for the GMM approximation of the likelihood ratio.  In the interest of space, results for the Bird, Bukin06 and Modified Rosenbrock functions have been relegated to the Supplementary Material.

\begin{figure}[!ht]
\centering 
\subfloat[Ackley]{\includegraphics[width=6.25in]{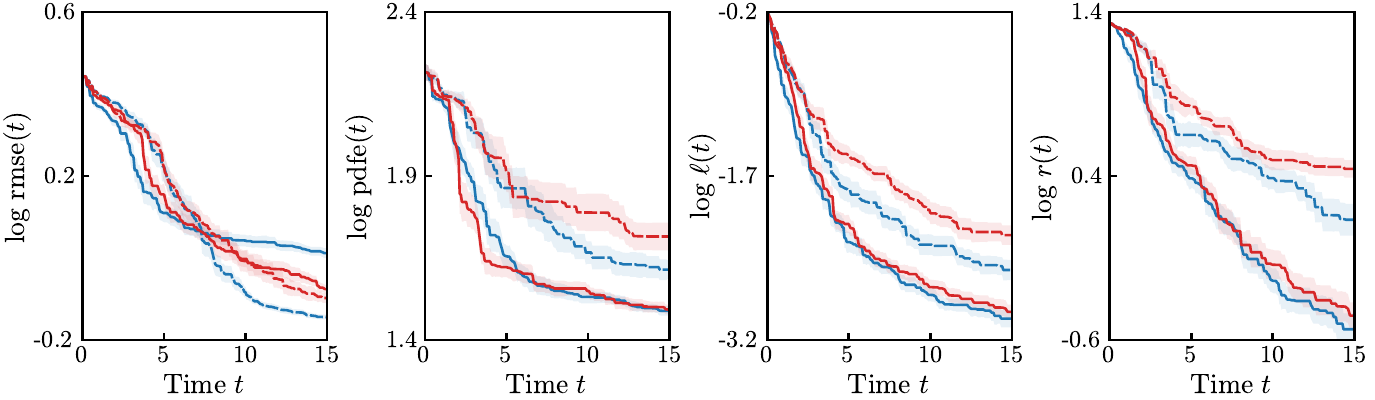}}

\subfloat[Michalewicz]{\includegraphics[width=6.25in]{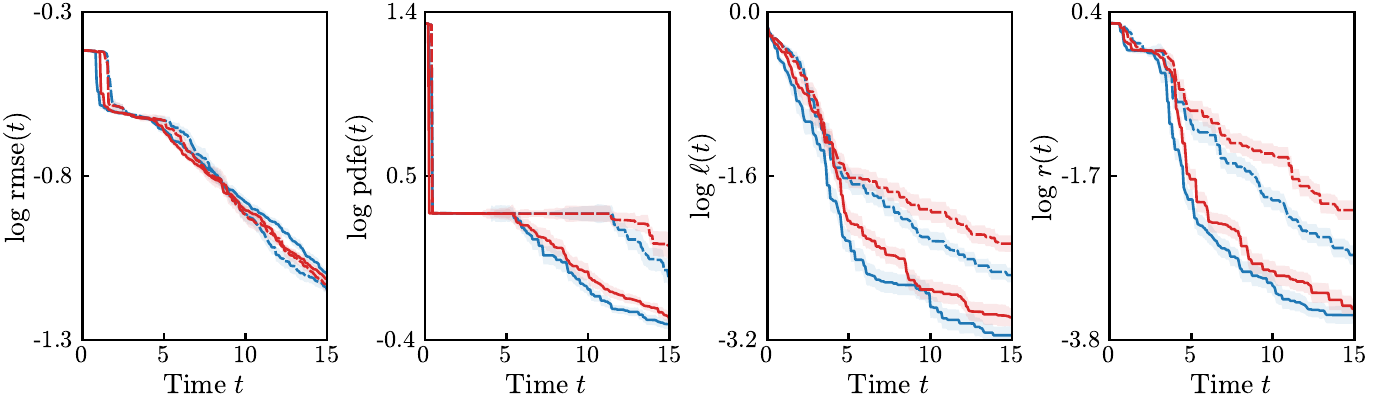}}

\caption{For uniform $p_\mathbf{z}$, performance of \bluedashedline~US; \bluesolidline~US-LW; \reddashedline~IVR; \redsolidline~IVR-LW.}
\label{fig:4}
\end{figure}

We first consider the situation in which the UAV operator has no prior beliefs about the locations of anomalies in the search space, and thus a uniform prior is used for $p_\mathbf{z}$.  For the Ackley and Michalewicz functions, Figure \ref{fig:4} shows that the proposed LW criteria substantially outperform their unweighted counterparts in the three critical metrics $\text{pdfe}$, $\ell$ and $r$, often by more than one full order of magnitude.  This shows that an UAV guided by US-LW or IVR-LW is able to identify environment anomalies more quickly and more efficiently than otherwise (Movies 1a and 1b).  When performance is measured in terms of the rmse, US-LW and IVR-LW are on par with US and IVR.  This is not surprising since US and IVR are specifically designed for rmse minimization, while US-LW and IVR-LW are not.  This is inconsequential from the standpoint of anomaly detection since the rmse only accounts for second-order moments and therefore is not a good indicator of anomalies.  Similar trends are seen for the other test functions (Figure S1 and Movies 1c--1e).

\begin{figure}[!ht]
\centering 
\subfloat[\label{fig5a}Ackley]{\includegraphics[width=6.25in]{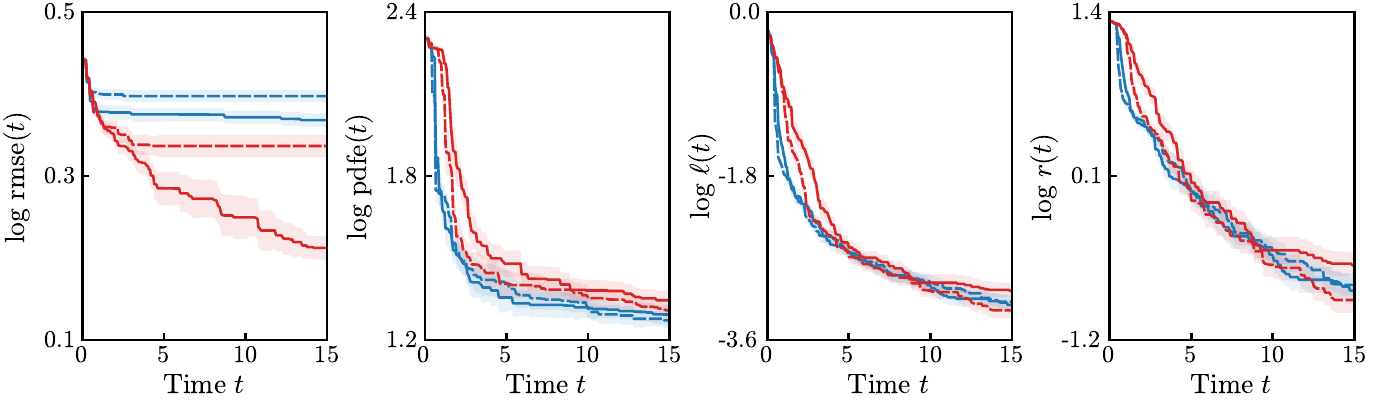}}

\subfloat[\label{fig5b}Michalewicz]{\includegraphics[width=6.25in]{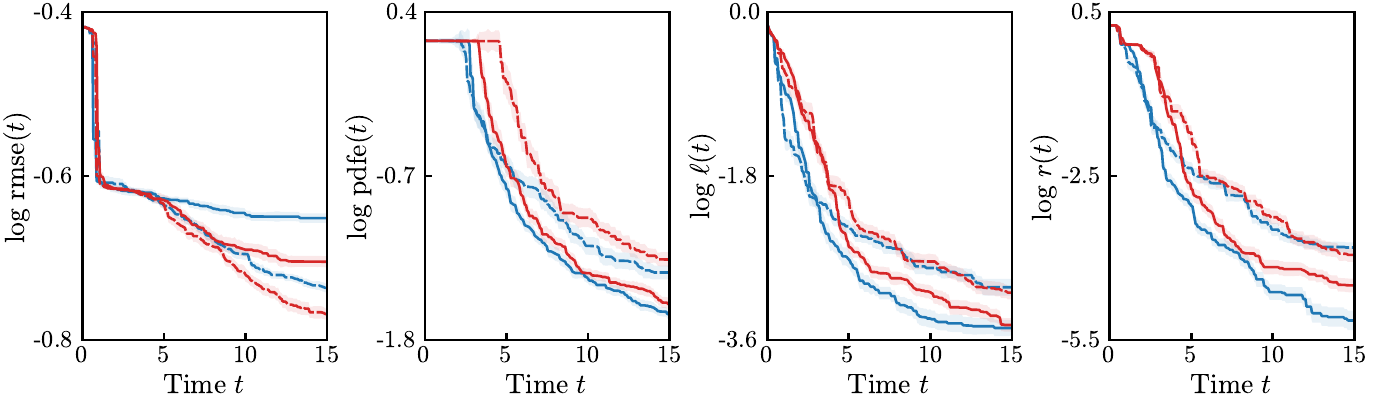}}

\caption{For Gaussian $p_\mathbf{z}$, performance of \bluedashedline~US-IW; \bluesolidline~US-LW; \reddashedline~IVR-IW; \redsolidline~IVR-LW.}
\label{fig:5}
\end{figure}

We next consider the situation in which the UAV operator has some prior beliefs about where anomalies may be located and therefore decides to specify the same spatial prior as in Figure \ref{fig:1} to focus exploration in the central part of the domain.  For the Ackley function, this is a good guess, and as a result US-IW, US-LW, IVR-IW, and IVR-LW deliver similar performance, as shown in Figure \ref{fig5a} and Movie 2a.  For this function, the likelihood ratio is not particularly helpful.  For the other test functions, however, the operator guess is quite poor, which leads to vastly different outcomes.  Figures \ref{fig5b} and S2 as well as Movies 2b--2e are a testament to the likelihood ratio ability to correct the operator beliefs and refine the UAV decision making in a way that makes anomaly detection more efficient.  Absent the likelihood ratio, the UAV does not have the ability to override the operator guidelines, missing out on the critical anomalous features that lie beyond its ascribed area.

In the results presented above, no information was provided to the algorithm regarding the fact that the environment did not depend on time.  Consequently, the UAV had to figure this out on its own from the data it had collected.  To let the UAV know that the environment is indeed static (as is the case with, for example, topography), the UAV operator may either set the lengthscale associated with the temporal variable in \eqref{eq:3} to infinity, or drop the time variable from the GP model altogether.  For the Ackley and Michalewicz functions with uniform prior, Figures S3 and S4 show that both approaches produce results that are quantitatively similar to those in Figures \ref{fig:4}  and \ref{fig:5} where time-independence was not explicitly enforced.  This demonstrates the ability of the spatiotemporal GP model to infer the critical characteristics of the environment.

\subsection{Benchmark Results for Dynamic Test Functions}
\label{sec:43}

\begin{figure}[!ht]
\centering \setcounter{subfigure}{-2}
\subfloat{\includegraphics[width=1.2in]{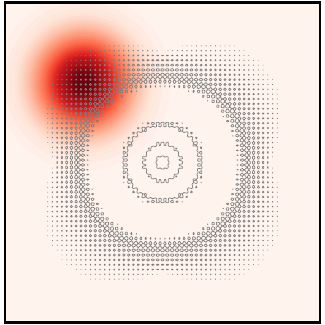}} ~
\subfloat{\includegraphics[width=1.2in]{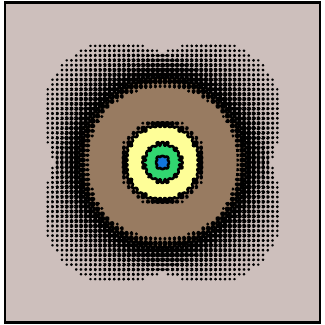}} ~ 
\subfloat[Ackley]{\includegraphics[width=1.2in]{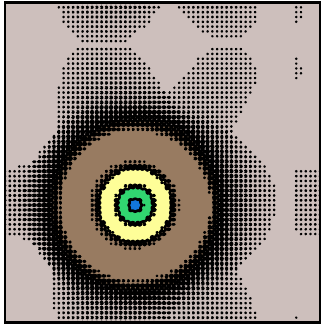}} ~
\subfloat{\includegraphics[width=1.2in]{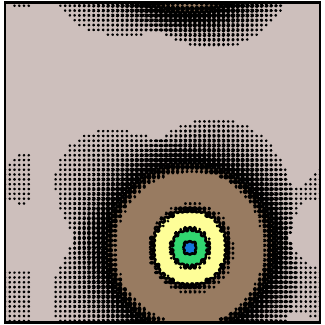}} ~
\subfloat{\includegraphics[width=1.2in]{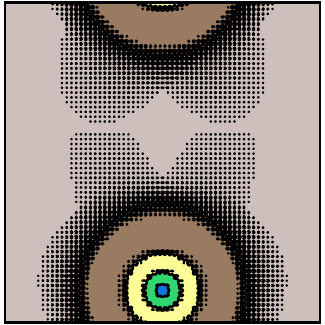}} 

\setcounter{subfigure}{-1}
\subfloat{\includegraphics[width=1.2in]{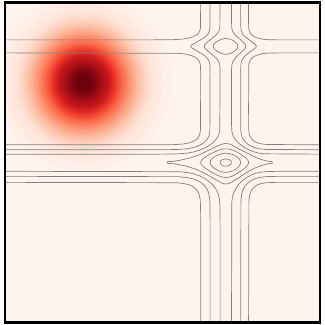}} ~
\subfloat{\includegraphics[width=1.2in]{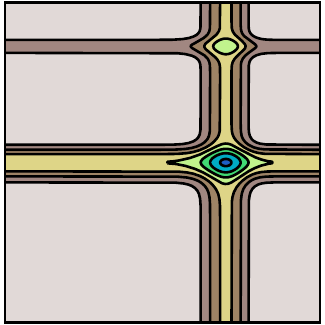}} ~
\subfloat[Michalewicz]{\includegraphics[width=1.2in]{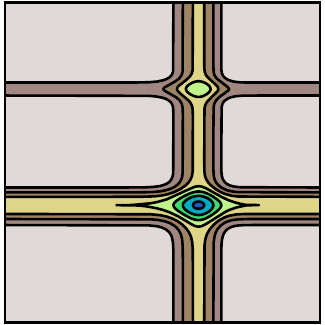}} ~
\subfloat{\includegraphics[width=1.2in]{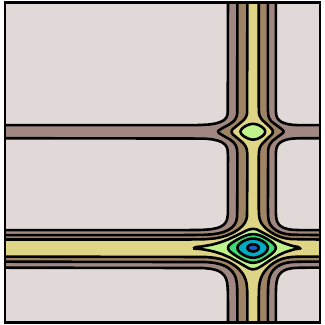}} ~
\subfloat{\includegraphics[width=1.2in]{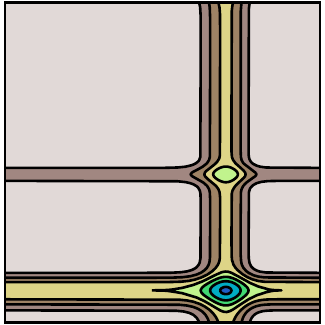}} 

\caption{From left to right: Contour plots of the adversarial input prior $p_\mathbf{z}$ (leftmost panel) and the map $f(\mathbf{z},t)$ at $t=0$, 5, 10 and 15 (four rightmost panels).}
\label{fig:6}
\end{figure}

We now investigate the performance of the algorithm under dynamic environmental conditions.  We repeat the experiments in Section \ref{sec:42} for the dynamic Ackley and Michalewicz functions which are constructed by applying the transformation 
\begin{gather*}
z_1 \longrightarrow z_1 + 0.1 \sin(2 \pi t/15) \mod 1\\
z_2 \longrightarrow 0.4 t / 15 \mod 1
\end{gather*}
to the physical space $[0,1]^2$.  (The modulo operator ensures periodicity of the resulting functions across the domain boundaries.)  We use a strongly adversarial Gaussian prior, $p_\mathbf{z} = \mathcal{N}([0.25,0.75]^\mathsf{T}, 0.01\mathbf{I})$, whose mass is primarily distributed in the upper left corner of the domain, quite far from any noteworthy environmental feature (see Figure \ref{fig:6}).

\begin{figure}[!ht]
\centering
\subfloat[\label{fig7a}Ackley]{\includegraphics[width=6.25in]{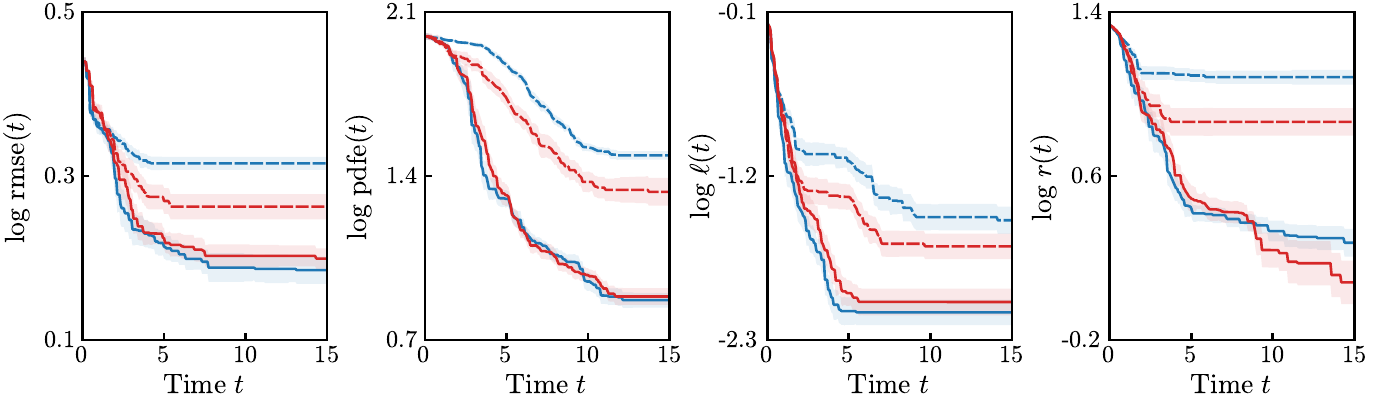}}

\subfloat[\label{fig7b}Michalewicz]{\includegraphics[width=6.25in]{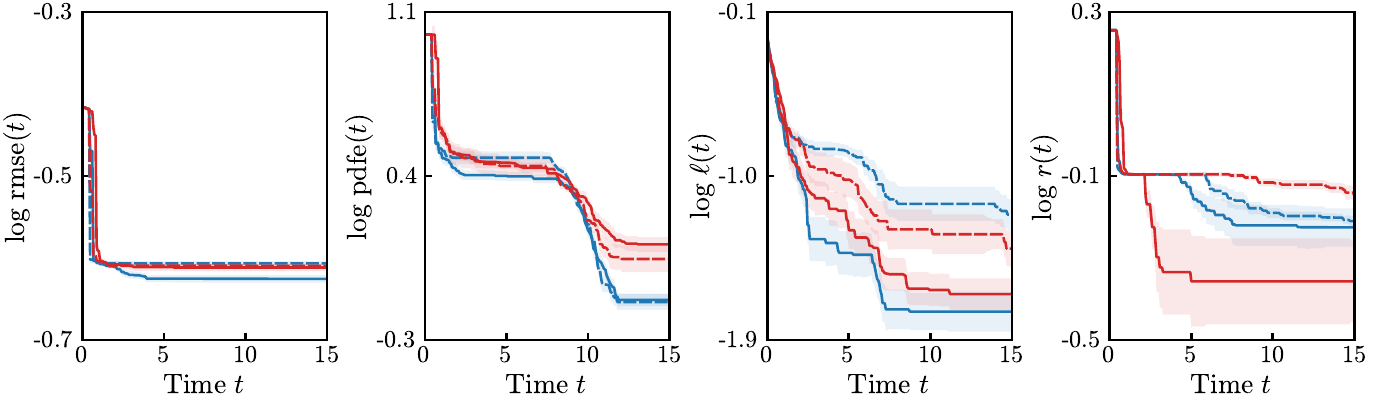}}
\caption{For the dynamic test functions with strongly adversarial prior shown in Figure \ref{fig:6}, performance of \bluedashedline~US-IW; \bluesolidline~US-LW; \reddashedline~IVR-IW; \redsolidline~IVR-LW.}
\label{fig:7}
\end{figure}

For the same parameters as in Section \ref{sec:42} ($\sigma_n^2=10^{-3}$, $L=0.2$, $\alpha=3\pi/4$, $R=0.02$, and $n_\textit{GMM}=2$), Figure \ref{fig:7} show that the presence of the likelihood ratio significantly improves algorithm performance despite the added difficulty arising from the environment being dynamic and the input prior being severely misleading (see also Movies 3a and 3b). These results illustrate the utility of the likelihood ratio for dynamic tracking of anomalies in strongly adversarial conditions.  We note, however, that the tracking of extremes cannot be perfect since the drone operates with limited resources and therefore must find a delicate balance between exploring the space, keeping track of the anomalies, and trusting or overcoming the prior.

\subsection{Application to Real-World Bathymetry Data}
\label{sec:44}

\begin{figure}[!ht]
\centering 
\includegraphics[width=3in]{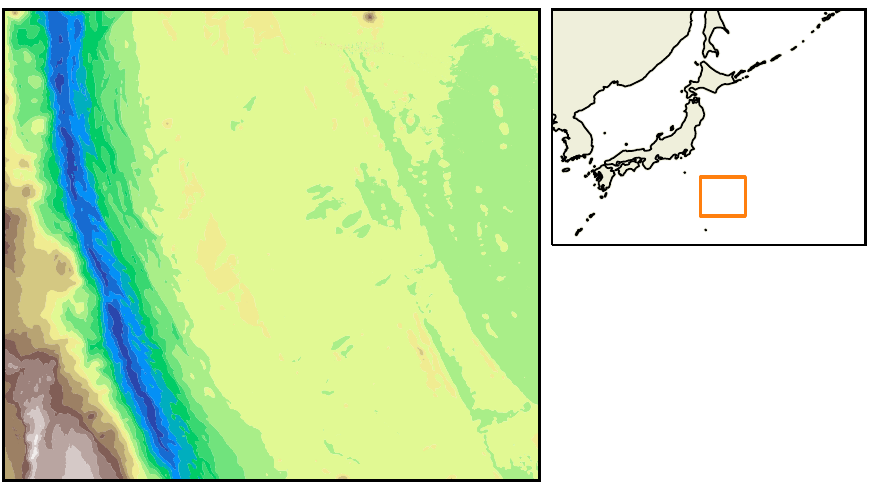}
\caption{Contour plot of the Izu--Ogasawara bathymetric profile located off the coast of Japan.}
\label{fig:cart}
\end{figure}

Bathymetric anomalies are important for the telecommunications industry and the oil and gas industry to be able to build subsea infrastructure, and also for navigation and coastal management as well as tsunami forecasting \cite{tani2017understanding}.  Here we consider the problem of reconstructing the topography of the seafloor near the Izu--Ogasawara trench, an oceanic trench located in the western Pacific Ocean and stretching from Japan to the northernmost section of the Mariana Trench.  The UAV is directed to explore an area of about 92,000 square miles (about 240,000 square kilometers) which is mostly flat, except for the trench itself whose deepest point is at about 32,000 ft (about 9,800 meters) below sea level.  The trench, therefore, constitutes a strong anomaly relative to the rest of the environment (Figure \ref{fig:cart}).

The true map $f$ is constructed by fitting a cubic spline to gridded bathymetric data made available by GEBCO\footnote{\url{https://www.gebco.net}}.   Since the dataset consists of real-world measurements, there is no need to corrupt it with artificial noise; we just let the UAV learn the value of the noise variance $\sigma_n^2$ from the raw data recorded during the mission.  Because the search space is rescaled to the unit square (see Section \ref{sec:41}), we use the same parameters for the path-planning algorithm as in the previous sections ($L=0.2$, $\alpha=3\pi/4$, and $R=0.02$).  For the GMM approximation of the likelihood ratio, we use $n_\textit{GMM}=2$.

\begin{figure}[!ht]
\centering 
\subfloat[\label{fig8a}Uniform $p_\mathbf{z}$]{\includegraphics[width=6.25in]{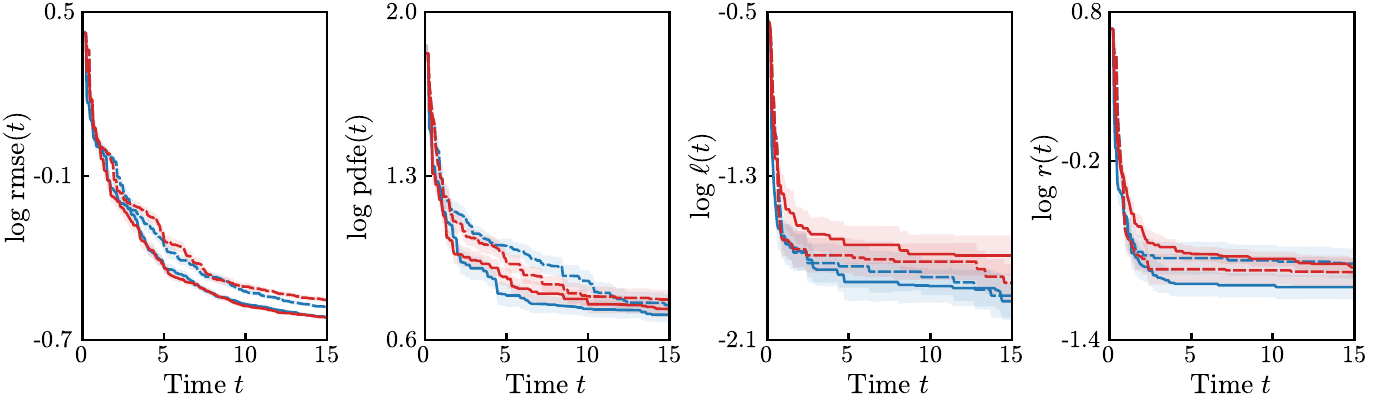}}

\subfloat[\label{fig8b}Gaussian $p_\mathbf{z}$]{\includegraphics[width=6.25in]{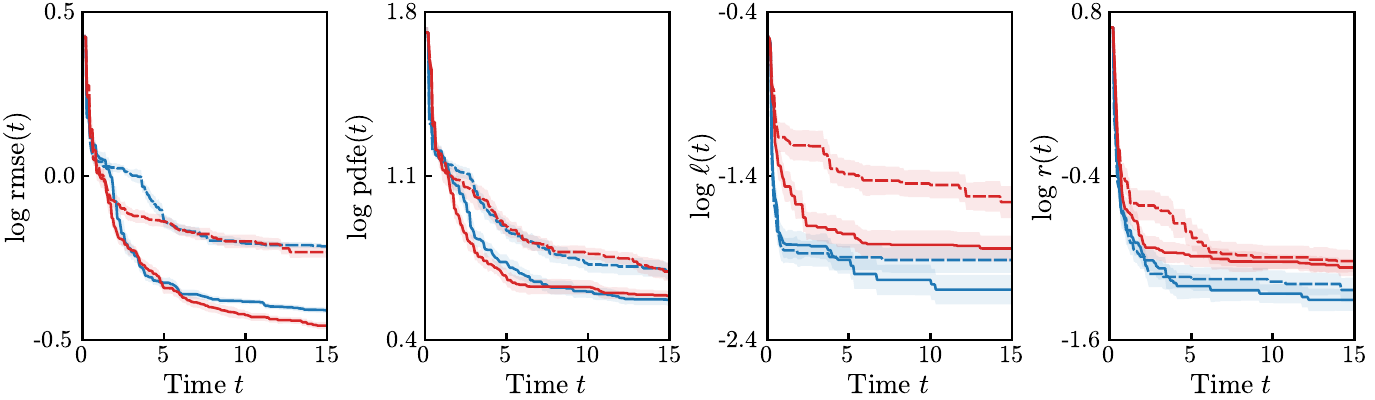}}

\caption{For the Izu--Ogasawara bathymetric profile, performance of \bluedashedline~US(-IW); \bluesolidline~US-LW; \reddashedline~IVR(-IW); \redsolidline~IVR-LW.}
\label{fig:8}
\end{figure}

Figure \ref{fig8a} shows that when no hint is given to the UAV prior to the mission (i.e., uniform $p_\mathbf{z}$), the likelihood-weighted acquisition functions lead to marginally better performance compared to the unweighted criteria.  On the other hand, figure \ref{fig8b} shows that when the UAV is given inaccurate prior information about the location of the trench (i.e., same Gaussian $p_\mathbf{z}$ as in Figure \ref{fig:1}), the likelihood-weighted acquisition functions perform substantially better than the unweighted criteria.  These results illustrate the power of the likelihood ratio to overcome adversarial beliefs, which is particularly valuable in situations where a malicious agent attempts to seize control of the UAV and lead it astray.

\section{Conclusions}
\label{sec:5}

We have introduced two novel acquisitions functions for informative path planning that are specifically designed for identification of anomalies in exploration missions.  Having ramifications running deep within the field of rare-event quantification, the proposed criteria exploit the unique properties of the likelihood ratio to guide the UAV towards regions of the space in which the quantity of interest is thought to exhibit strong anomalies.  We applied the proposed algorithm to a number of synthetic test functions as well as a real-world bathymetry dataset.  Overall, the likelihood-weighted criteria led to faster identification of anomalies present in the environment, especially in adversarial settings.  

Our approach can be adapted for use with more sophisticated path-planning algorithms and in more complex environments.  The two-dimensional Dubins path parametrization used in this work can be extended to a three-dimensional search space using the implementation of \cite{mclain2014implementing}.  As discussed in Section \ref{sec:22}, more complex time-dependence of the environment can be accommodated by using a covariance function that encodes specific spatiotemporal relationships \cite{singh2010modeling,krause2011contextual,marchant2014bayesian}.  This is important from the standpoint of environment surveillance.  Localization noise can also be accounted for using the algorithm of \cite{oliveira2019bayesian,oliveira2020bayesian}.

Our method is quite general and therefore has positive implications for a range of commercial, scientific, and military applications in which anomaly detection is critical.  For underwater UAVs, our approach can be used to make detailed maps of the seafloor, of which less is known than of the topography of Mars \cite{tani2017understanding}.  In the Arctic region, bathymetry can be used to map the fjords' sill depths, a key indicator for the rise in global sea level.  Our method can also improve detection of wreckages of missing crafts as well as other artifacts found in underwater archeological sites, which may be viewed as abnormal features lying on the ocean floor.  

For aerial UAVs, our work can be used to detect abnormal concentrations of chemicals, pollutants, or radioactive material in a region of interest; to reconstruct terrain with unknown topography in order to plan large-scale architecture; and to monitor abnormal crop growth in order to optimize agriculture operations.  In the military, our algorithm can facilitate the clearing of minefields, having the ability to quickly identify signs of explosive chemicals leaking from landmines and causing abnormalities in the surrounding vegetation \cite{bristol2016bristol}.  
\section*{Acknowledgments}
The authors acknowledge support from the Air Force Office of Scientific Research (MURI Grant No. FA9550-21-1-0058), the Army Research Office (Grant No. W911NF-17-1-0306) and the 2020 MathWorks Faculty Research Innovation Fellowship.

\section*{References}
\bibliographystyle{jfm}
\bibliography{bibl}

\end{document}


\begin{frontmatter}

\title{Informative Path Planning for Extreme Anomaly Detection in \\ Environment Exploration and Monitoring: \\ Supplementary Material} 

\author[]{Antoine Blanchard}
\author[]{Themistoklis Sapsis}

\end{frontmatter}

\section{Equivalence Between US, UCB, PI, and EI in the Limit of Pure Exploration}
\label{app:1}

\subsection{Preliminaries}

We begin with a few definitions.  To avoid any ambiguity, we follow the convention that the acquisition functions are to be maximized (i.e., UCB, PI and EI are looking to \emph{maximize} the function $f$).  

\paragraph{Uncertainty sampling (US):} 
\begin{equation}
a_\textit{US}(\mathbf{x}) = \sigma^2(\mathbf{x})
\label{eq:S1}
\end{equation}

\paragraph{Upper confidence bound (UCB):}  
\begin{equation}
a_\textit{UCB}(\mathbf{x}) = \mu(\mathbf{x}) + \kappa \sigma(\mathbf{x})
\label{eq:S2}
\end{equation}

\paragraph{Probability of improvement (PI):} 
\begin{equation}
a_\textit{PI}(\mathbf{x}) =\Phi (\lambda(\mathbf{x}))
\label{eq:S3}
\end{equation}

\paragraph{Expected improvement (EI):} 
\begin{equation}
a_\textit{EI}(\mathbf{x}) = \sigma(\mathbf{x}) \left[\lambda(\mathbf{x})\Phi(\lambda(\mathbf{x})) + \phi(\lambda(\mathbf{x})) \right]
\label{eq:S4}
\end{equation}

In the above, $\Phi$ and $\phi$ are the cumulative and probability density functions of the standard normal distribution, respectively, and $\kappa$ is a positive parameter that controls the trade-off between exploration (large $\kappa$) and exploitation (small $\kappa$). We have also introduced the quantity
\begin{equation}
\lambda(\mathbf{x}) = \frac{\mu(\mathbf{x}) - y^*  - \kappa}{\sigma(\mathbf{x})},
\end{equation}
where $y^*$ denotes the current best observation.  

We first note that the locations and ordering of a function's extrema are preserved under strictly increasing continuous transformations. Therefore, to prove that UCB, PI and EI are equivalent to US in the limit of pure exploration (i.e., large $\kappa$), we only need to show that in that limit, UCB, PI and EI can be expressed as strictly increasing continuous functions of US.  We will use the fact that the composition of two strictly increasing functions is also a strictly increasing function.


\subsection{Equivalence Between UCB and US in the Limit of Pure Exploration}

It is straightforward to see that for large $\kappa$,
\begin{equation}
a_\textit{UCB}(\mathbf{x}) \sim \kappa \sigma(\mathbf{x}).
\label{eq:S7}
\end{equation}
The function $x \longmapsto x^2$ is strictly increasing on $[0, +\infty)$, which completes the proof.

\subsection{Equivalence between PI and US in the Limit of Pure Exploration}

For large $\kappa$, we note that
\begin{equation}
a_\textit{PI}(\mathbf{x}) \sim \frac{1}{2}\left\{1 - \mathrm{erf}\!\left[\frac{\kappa}{\sqrt{2}\sigma(\mathbf{x})}\right]\right\} \!,
\label{eq:S9}%
\end{equation}
where we have used the fact that $\Phi(-x) = (1/2)[1 - \mathrm{erf}(x/\sqrt{2})]$ for any $x$.  For any $a>0$, the function $x \longmapsto 1-\mathrm{erf}(a/\sqrt{x})$ is strictly increasing on $[0, +\infty)$, which completes the proof.

\subsection{Equivalence between EI and US in the Limit of Pure Exploration}
For large $\kappa$, we have 
\begin{equation}
a_\textit{EI}(\mathbf{x}) \sim \frac{\sigma(\mathbf{x})}{\sqrt{2\pi}} \exp\!\left[-\frac{\kappa^2}{2\sigma^2(\mathbf{x})}\right] -\frac{\kappa}{2}\left\{1 - \mathrm{erf}\!\left[\frac{\kappa}{\sqrt{2}\sigma(\mathbf{x})}\right]\right\}\!. \label{eq:S13}
\end{equation}
But for $x \gg 1$, we know that  
\begin{equation}
\mathrm{erf}(x) \sim 1 - \frac{\exp(-x^2)}{x \sqrt{\pi}} \left[ 1 - \frac{1}{2x^2} \right]\!.
\end{equation}
It follows that  
\begin{equation}
a_\textit{EI}(\mathbf{x}) \sim \frac{\sigma(\mathbf{x})}{\sqrt{2\pi}} \exp\!\left[-\frac{\kappa^2}{2\sigma^2(\mathbf{x})}\right] -  \frac{\sigma(\mathbf{x})}{\sqrt{2\pi}} \exp\!\left[-\frac{\kappa^2}{2\sigma^2(\mathbf{x})}\right]  \left[ 1 - \frac{\sigma^2(\mathbf{x})}{\kappa^2} \right]\!,
\end{equation}
and therefore
\begin{equation}
a_\textit{EI}(\mathbf{x}) \sim  \frac{\sigma^3(\mathbf{x})}{\kappa^2\sqrt{2\pi}} \exp\!\left[-\frac{\kappa^2}{2\sigma^2(\mathbf{x})}\right]\!.
\label{eq:S16}
\end{equation}
For any $a>0$, the function $x \longmapsto x^{3/2} \exp(-a/x)$ is strictly increasing on $[0, +\infty)$, which completes the proof.

\section{Analytical Expressions for Benchmark Test Functions}
\label{app:4}


\begin{subequations}
\paragraph{Ackley function:}
\begin{equation}
f(z_1,z_2) = -a \exp \! \left[ -b \sqrt{(z_1^2+z_2^2)/2} \right] - \exp\! \left[(\cos{cz_1}+\cos{cz_2})/2 \right] +a +\exp(1),
\end{equation}
where $a=20$, $b=0.2$, and $c=2\pi$.  

\paragraph{Bird function:}
\begin{equation}
f(z_1,z_2) = \sin(z_1)\exp[(1-\cos z_2)^2] + \cos(z_2)\exp[(1-\sin z_1)^2] + (z_1-z_2)^2
\end{equation}

\paragraph{Bukin function:}
\begin{equation}
f(z_1,z_2) = 100 \sqrt{|z_2-0.01z_1^2|} + 0.01 |z_1+10|
\end{equation}

\paragraph{Michalewicz function:}
\begin{equation}
f(z_1,z_2) = -  \sin(z_1) \sin^{2m}(z_1^2/\pi) -  \sin(z_2) \sin^{2m}(2 z_2^2/\pi),
\label{eq:micha}
\end{equation}
where $m$ controls the steepness of the valleys and ridges.  In this work we use $m=10$.  

\paragraph{Modified Rosenbrock function:}
\begin{equation}
f(z_1,z_2) = 74 + 100 (z_2-z_1^2)^2 + (1-z_1)^2 - 400 \exp [-10(z_1+1)^2-10(z_2+1)^2]
\end{equation}

\end{subequations}

\clearpage
\begin{figure}[p]
\centering 
\subfloat[Bird]{\includegraphics[width=6.25in]{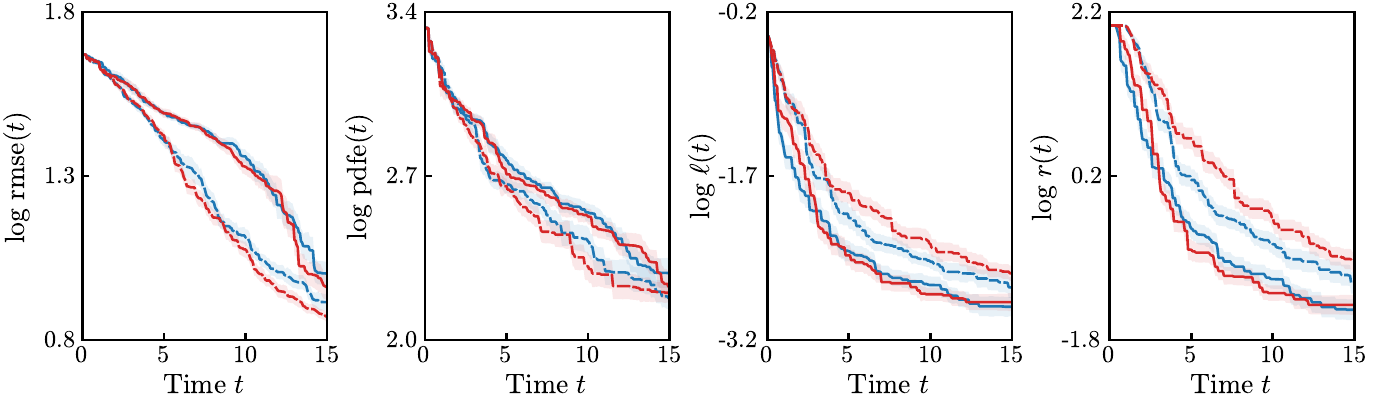}}

\subfloat[Bukin06]{\includegraphics[width=6.25in]{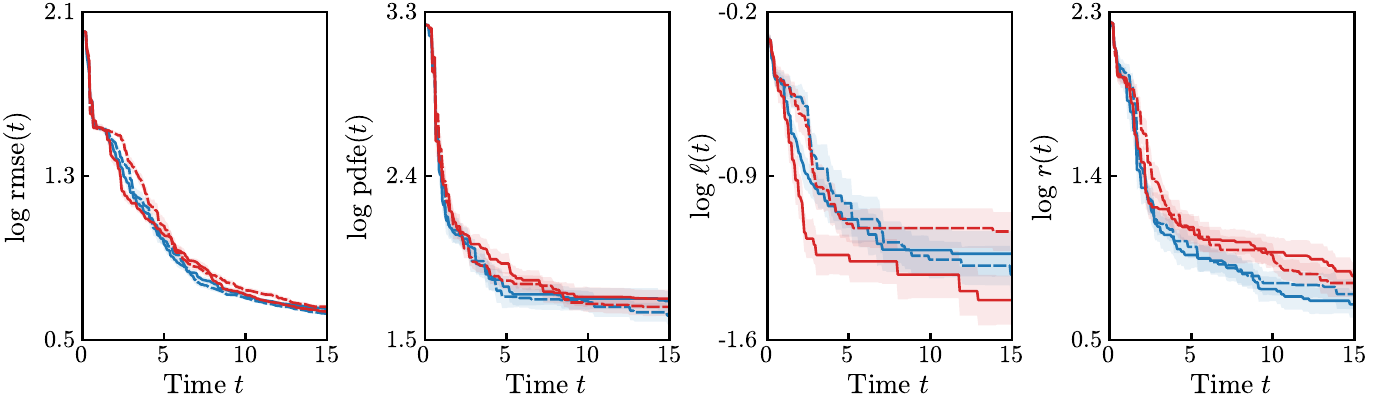}}

\subfloat[Modified Rosenbrock]{\includegraphics[width=6in]{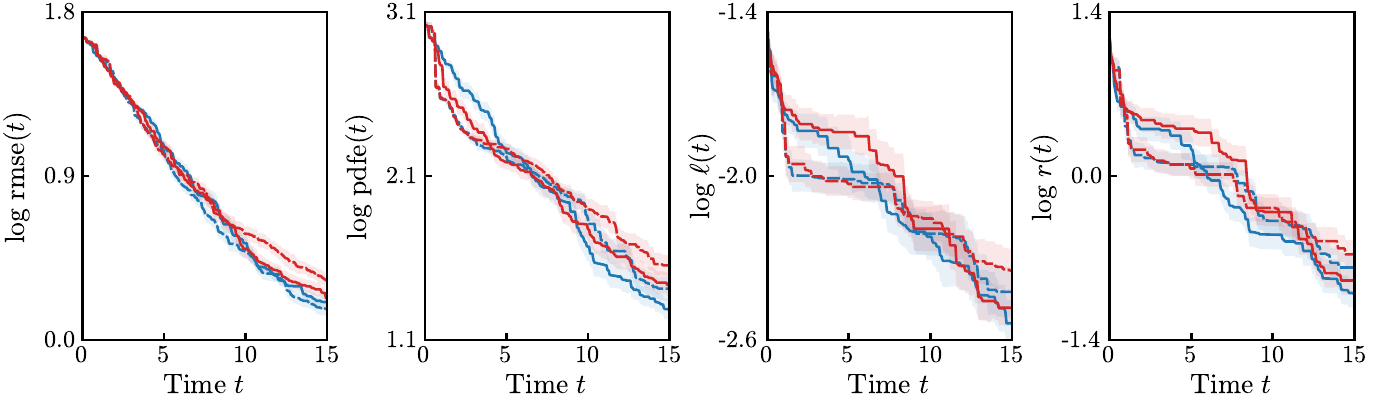}}

\caption{For uniform $p_\mathbf{z}$, performance of \bluedashedline~US; \bluesolidline~US-LW; \reddashedline~IVR; \redsolidline~IVR-LW.}
\label{fig:S1}
\end{figure}

\clearpage
\begin{figure}[p]
\centering 
\subfloat[Bird]{\includegraphics[width=6.25in]{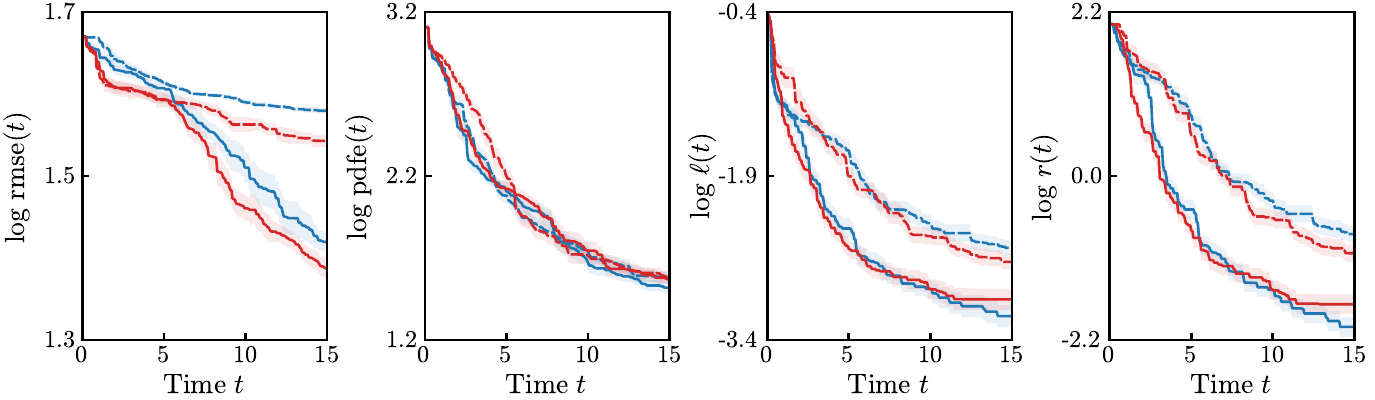}}

\subfloat[Bukin06]{\includegraphics[width=6.25in]{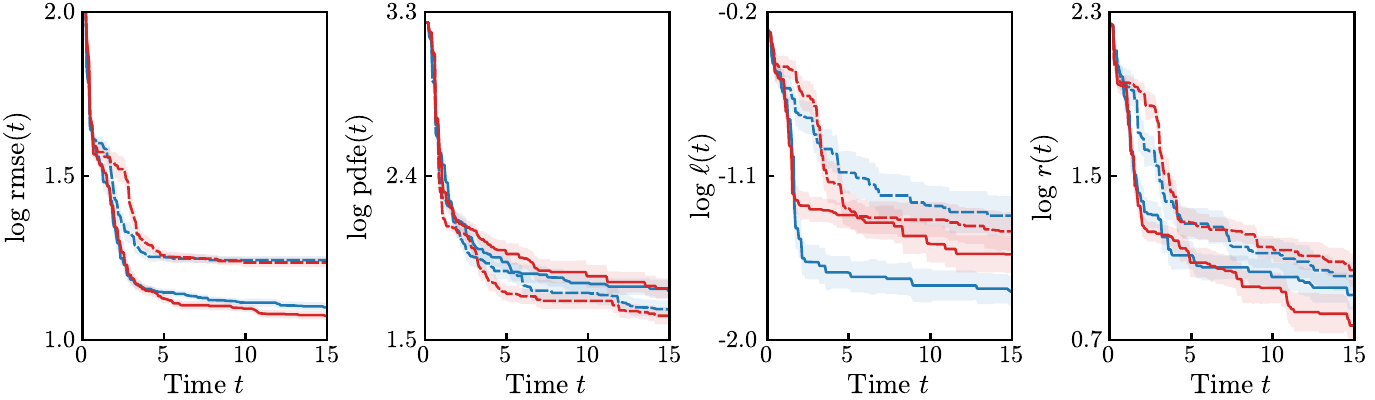}}

\subfloat[Modified Rosenbrock]{\includegraphics[width=6.25in]{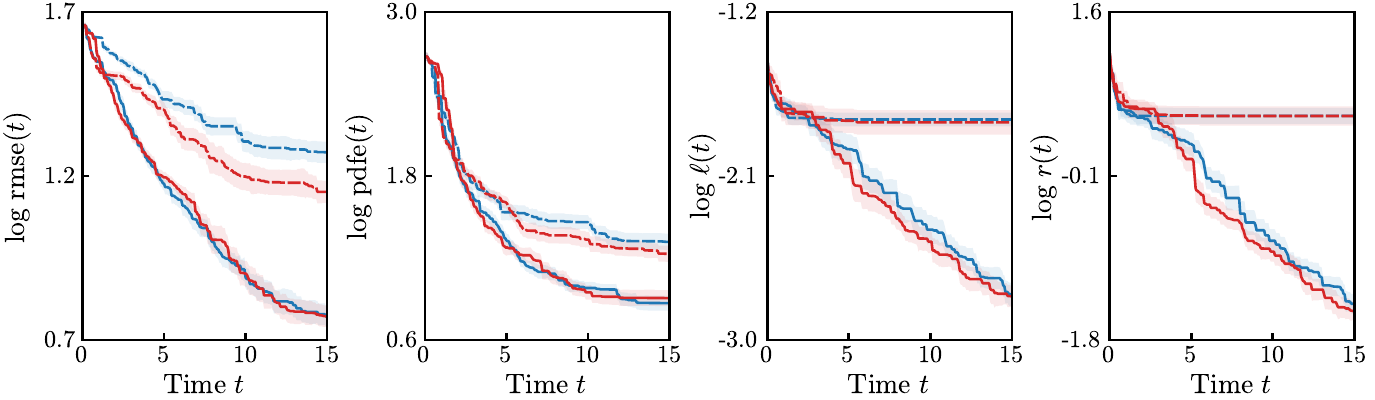}}

\caption{For Gaussian $p_\mathbf{z}$, performance of \bluedashedline~US-IW; \bluesolidline~US-LW; \reddashedline~IVR-IW; \redsolidline~IVR-LW.}
\label{fig:S2}
\end{figure}

\clearpage
\begin{figure}[p]
\centering 
\subfloat[Time dependence inferred from data (same as Figure 4a in main text)]{\includegraphics[width=6.25in]{figs/fig4a}}

\subfloat[Temporal lengthscale set to infinity in RBF kernel]{\includegraphics[width=6.25in]{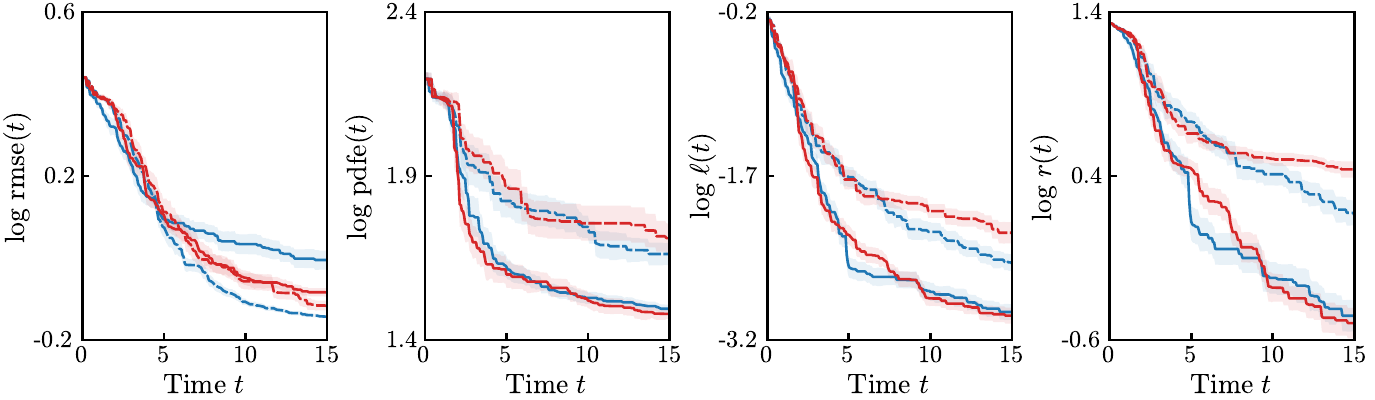}}

\subfloat[No temporal variable in GP model]{\includegraphics[width=6.25in]{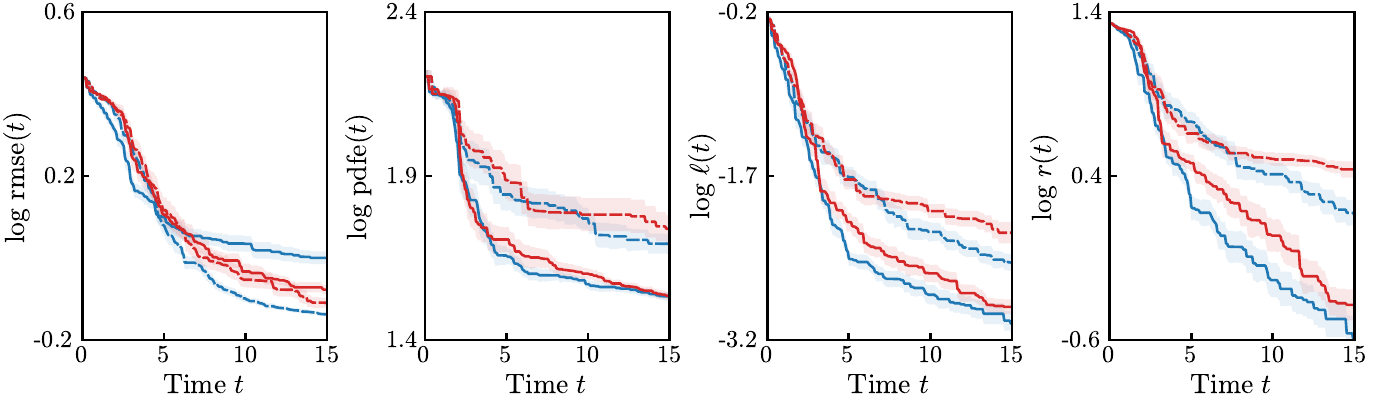}}

\caption{For the Ackley function with uniform $p_\mathbf{z}$, performance of \bluedashedline~US-IW; \bluesolidline~US-LW; \reddashedline~IVR-IW; \redsolidline~IVR-LW.}
\label{fig:S4}
\end{figure}

\clearpage
\begin{figure}[p]
\centering 
\subfloat[Time dependence inferred from data (same as Figure 4b in main text)]{\includegraphics[width=6.25in]{figs/fig4b}}

\subfloat[Temporal lengthscale set to infinity in RBF kernel]{\includegraphics[width=6.25in]{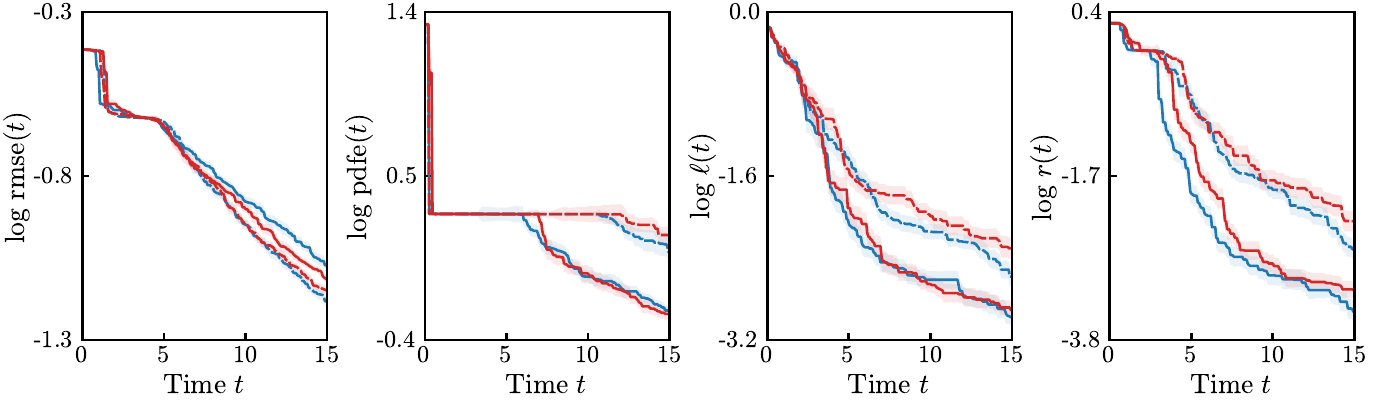}}

\subfloat[No temporal variable in GP model]{\includegraphics[width=6.25in]{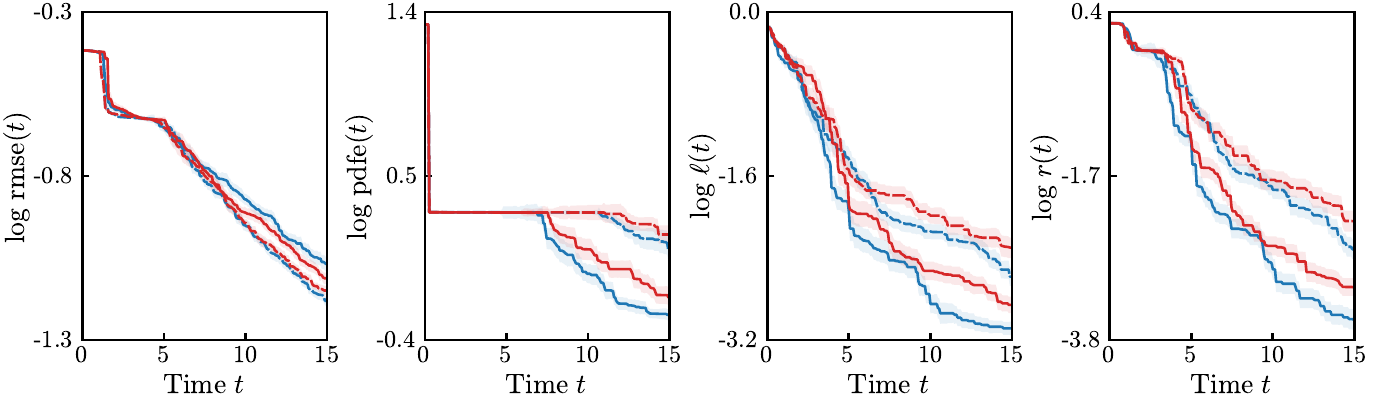}}

\caption{For the Michalewicz function with uniform $p_\mathbf{z}$, performance of \bluedashedline~US-IW; \bluesolidline~US-LW; \reddashedline~IVR-IW; \redsolidline~IVR-LW.}
\label{fig:S4}
\end{figure}

\clearpage

\section*{Movie captions}

\paragraph{Movie 1a:}  Comparison of IVR (pink) and IVR-LW (dark green) for the Ackley function with uniform $p_\mathbf{z}$ and the parameters given in Section 4.1.  
\paragraph{Movie 1b:}  Comparison of IVR (pink) and IVR-LW (dark green) for the Michalewicz function with uniform $p_\mathbf{z}$ and the parameters given in Section 4.1.
\paragraph{Movie 1c:}  Comparison of IVR (pink) and IVR-LW (dark green) for the Bird function with uniform $p_\mathbf{z}$ and the parameters given in Section 4.1.
\paragraph{Movie 1d:}  Comparison of IVR (pink) and IVR-LW (dark green) for the Bukin06 function with uniform $p_\mathbf{z}$ and the parameters given in Section 4.1.
\paragraph{Movie 1e:}  Comparison of IVR (pink) and IVR-LW (dark green) for the Modified Rosenbrock  function with uniform $p_\mathbf{z}$ and the parameters given in Section 4.1.

\paragraph{Movie 2a:}  Comparison of IVR-IW (pink) and IVR-LW (dark green) for the Ackley function with Gaussian $p_\mathbf{z}$ and the parameters given in Section 4.1.
\paragraph{Movie 2b:}  Comparison of IVR-IW (pink) and IVR-LW (dark green) for the Michalewicz function with Gaussian $p_\mathbf{z}$ and the parameters given in Section 4.1.  
\paragraph{Movie 2c:}  Comparison of IVR-IW (pink) and IVR-LW (dark green) for the Bird function with Gaussian $p_\mathbf{z}$ and the parameters given in Section 4.1.
\paragraph{Movie 2d:}  Comparison of IVR-IW (pink) and IVR-LW (dark green) for the Bukin06 function with Gaussian $p_\mathbf{z}$ and the parameters given in Section 4.1.
\paragraph{Movie 2e:}  Comparison of IVR-IW (pink) and IVR-LW (dark green) for the Modified Rosenbrock  function with Gaussian $p_\mathbf{z}$ and the parameters given in Section 4.1.

\paragraph{Movie 3a:}  Comparison of IVR-IW (pink) and IVR-LW (dark green) for the dynamic Ackley function with strongly adversarial Gaussian $p_\mathbf{z}$ and the parameters given in Section 4.2.
\paragraph{Movie 3b:}  Comparison of IVR-IW (pink) and IVR-LW (dark green) for the dynamic Michalewicz function with strongly adversarial Gaussian $p_\mathbf{z}$ and the parameters given in Section 4.2.
